\documentclass[runningheads]{llncs}
\usepackage[T1]{fontenc}
\usepackage{amssymb}
\usepackage{amsmath}
\usepackage{multirow}
\usepackage{graphicx,verbatim}
\usepackage{amsmath}

\usepackage[svgnames]{xcolor}
\usepackage{color}
\usepackage[
    colorlinks=true,
    linkcolor=blue,
    citecolor=blue,
    urlcolor=blue
]{hyperref}
\begin{document}
\title{GL-LFGNN:A Global-Local Dual-branch Causal Graph Neural Network Based on Liang-Kleeman Information Flow for  EEG Emotion Recognition}
\titlerunning{GL-LFGNN}
%

\author{Ziyi Wang\inst{1} \and Dongyang Kuang\inst{1}}
\authorrunning{Wang et al.}
\institute{School of Mathematics (Zhuhai), Sun Yat-sen University, Zhuhai, China\\
\email{wangzy286@mail.sysu.edu.cn} \\
\email{kuangdy@mail.sysu.edu.cn}}

\maketitle              

\begin{abstract}
EEG-based emotion recognition holds significant promise for objective diagnosis of mood disorders. Graph neural networks (GNNs) have emerged as the dominant paradigm for modeling inter-channel dependencies in EEG, yet existing approaches rely on symmetric adjacency matrices derived from spatial proximity or functional correlations that fundamentally capture statistical associations rather than directed causal influences, which conflicts with the inherently asymmetric, causally-driven nature of neural information flow. To bridge this gap, we propose \textbf{GL-LFGNN}, a Global-Local Dual-branch Causal Graph Neural Network grounded in Liang-Kleeman information flow theory. Unlike Granger causality that merely assesses temporal precedence, our approach rigorously quantifies causal strength from a dynamical systems perspective, yielding neurophysiologically interpretable directed graphs. A dual-branch architecture further integrates whole-brain connectivity with region-specific processing aligned to established functional neuroanatomy. On the MEEG dataset, GL-LFGNN achieves \textbf{86.17\% (Arousal)} and \textbf{86.71\% (Valence)} accuracy with only \textbf{37K parameters}---approximately 10\% of the current state-of-the-art---demonstrating that principled causal modeling can simultaneously enhance interpretability, generalization, and computational efficiency. Code will be released.


\end{abstract}

\section{Introduction}
Depression, anxiety disorders, and autism spectrum disorder (ASD) pose significant global health challenges, where early detection is essential for effective intervention. Unlike self-report questionnaires susceptible to subjective bias, scalp electroencephalography (EEG) provides an objective and high-temporal resolution measure of neural activity. The proliferation of lightweight, consumer-grade EEG headsets further enables continuous at-home emotion monitoring for mental wellness tracking, stress management, and personalized interventions. This convergence of clinical need and wearable technology positions EEG-based affective computing as a transformative tool—bridging the gap between laboratory research and real-world applications in digital mental healthcare.

In recent years, with the rapid development of artificial intelligence, deep learning technology has made significant progress in the field of emotion recognition based on EEG. Early research mainly uses convolutional neural networks (CNNs)~\cite{li2016emotion,schirrmeister2017deep,lawhern2018eegnet} and recurrent neural networks (RNNs)~\cite{zhang2019strnn,shen2020eeg} for feature extraction. Although these models excel at processing Euclidean spatial data, CNNs and RNNs have difficulty capturing spatial associations between electrode channels and brain network properties due to the inherently non-Euclidean topology of EEG signals, thus limiting recognition accuracy. To overcome this limitation, graph neural networks (GNNs) show great potential for improvement by explicitly modeling topological relationships between EEG channels.

In graph-based EEG learning, channels serve as nodes and their interactions define edges. Existing adjacency matrices fall into three categories: (1) \textit{distance-based} methods using electrode spatial proximity~\cite{zhong2022rgnn,li2023fusiongraph,chen2024dualatt,ping2025kan}, e.g., RBF kernels or 3D coordinates; (2) \textit{correlation-based} methods measuring statistical dependencies via Pearson correlation~\cite{hou2024gcnsnet} or phase-locking value~\cite{wang2019plvgcn}; and (3) \textit{data-driven} approaches like DGCNN~\cite{song2020dgnn} that learn adjacency weights end-to-end. \textbf{Critically, all these methods produce symmetric matrices}, implicitly assuming bidirectional equivalence between brain regions---a premise contradicted by decades of neuroimaging evidence demonstrating hierarchical, directional information flow in emotional processing circuits~\cite{liang2021normalized}.

Recent multi-scale architectures~\cite{zhong2022rgnn,ding2024lggnet,xiao2024meeg,gu2026hierarchical} incorporate brain regionalization priors to capture local-global features, yet still rely on symmetric connectivity. To model directionality, some studies~\cite{ramakrishna2021causal,manomaisaowapak2022gc,kong2023causal} employ Granger causality for directed graph construction. However, Granger causality has two fundamental limitations: (1) it assesses \textit{temporal precedence} rather than true causation---confounded variables with different time lags can produce spurious causal links; (2) it requires stationarity assumptions often violated by non-stationary EEG dynamics during emotional state transitions. These shortcomings motivate our adoption of Liang-Kleeman information flow, which derives causality from \textit{entropy transfer in dynamical systems}, providing a rigorous, physics-grounded alternative. 

To address these gaps, we propose \textbf{GL-LFGNN} with three key contributions:
\begin{enumerate}
    \item \textbf{Principled causal graph construction:} We introduce Liang-Kleeman information flow---rooted in entropy dynamics of coupled systems---to construct directed adjacency matrices with rigorous causal semantics, moving beyond correlational proxies.
    \item \textbf{Neuroanatomy-informed dual-branch design:} A global branch captures whole-brain causal pathways while a local branch models intra-regional interactions aligned with functional specialization (frontal: regulation; temporal: auditory processing), enabling interpretable multi-scale fusion.
    \item \textbf{Efficiency without compromise:} GL-LFGNN achieves SOTA-comparable accuracy using only $\sim$10\% of the parameters of leading methods, demonstrating that explicit causal structure reduces reliance on model capacity.
\end{enumerate}

\section{Methods}
\begin{figure}
    \vspace{-10pt}
    \centering
    \includegraphics[width=\linewidth]{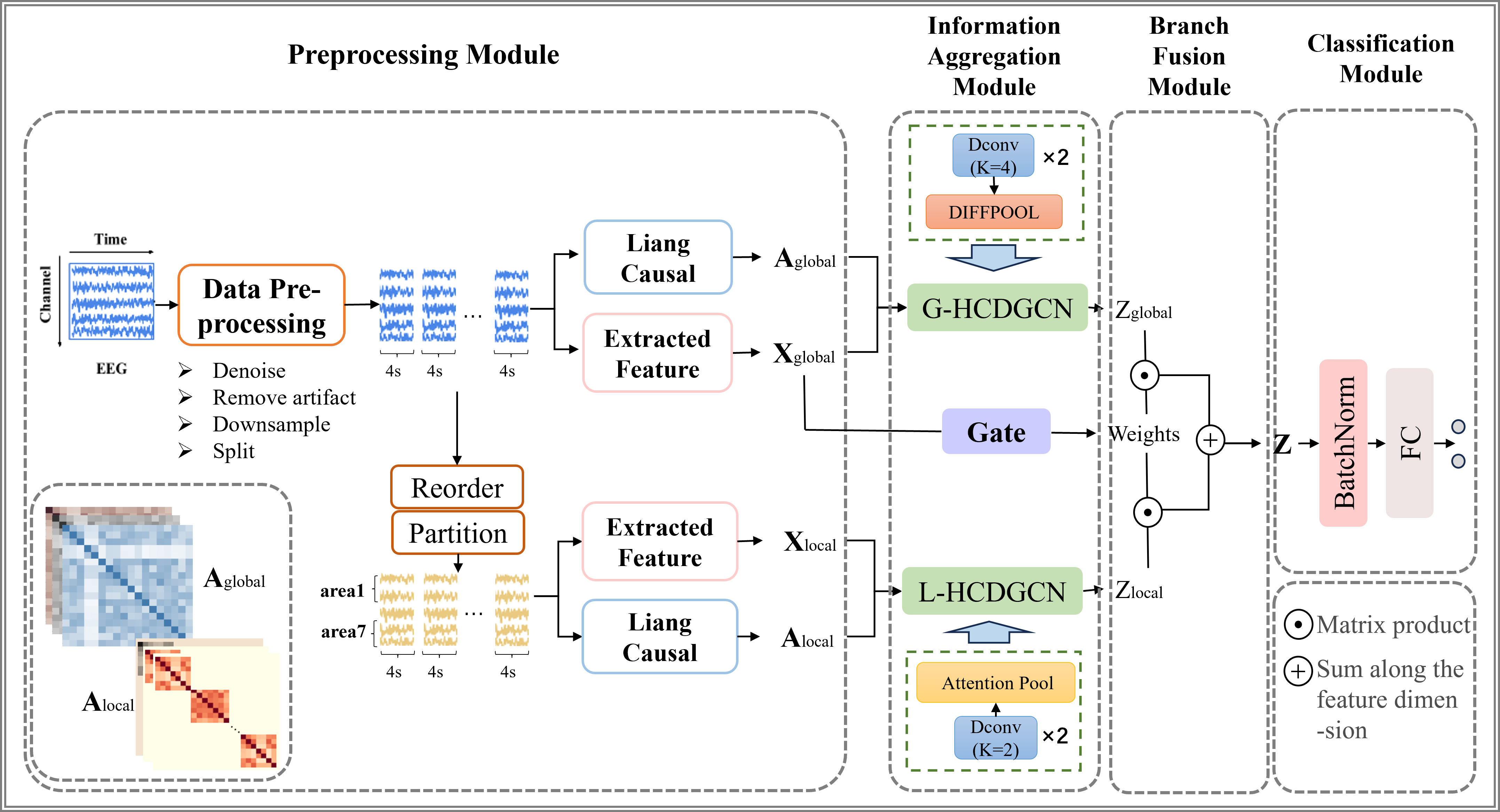}
    \caption{The overall framework of GL-LFGNN}
    \label{fig:1}
\end{figure}

The overall framework of the proposed model is shown in Fig.\ref{fig:1}. GL-LFGNN comprises four modules: preprocessing, information aggregation, branch fusion and classification.We will introduce Liang-Kleeman information flow theory, which is the method of causal graph construction, and then introduce the working principle of the four modules.

\vspace{-20pt}
\subsection{Causal graph construction}\label{sec:2.1}

Liang--Kleeman information flow quantifies directed information transfer between components of a dynamical system based on entropy variation, and provides an effective tool for characterizing causal interactions in multivariate time series~\cite{liang2021normalized}. 
Within this framework, a variable $X_j$ is considered causal to $X_i$ if the information flow $T_{j\to i}$ is non-zero, and the magnitude of causality is measured by $|T_{j\to i}|$.

In practice, the underlying dynamics of EEG signals are unknown. Under the assumption of a linear stochastic system, Liang~\cite{liang2021normalized} derived an estimator for the information flow from $X_j$ to $X_i$ in a $d$-dimensional system:

\vspace{-10pt}
\begin{align}
    \label{equ:1}
    \hat{T}_{j \to i} = \frac{1}{\det \mathbf{C}} \cdot \sum_{k=1}^{d} \Delta_{jk} C_{k,di} \cdot \frac{C_{ij}}{C_{ii}},
\end{align}
where $\mathbf{C}$ is the covariance matrix, $\Delta_{jk} $ is the algebraic cofactor of $\mathbf{C}$, and $C_{i,dj}=\overline{(X_{i}-\bar{X}_{i})(\dot{X}_{j}-\overline{\dot{X}_{j}})}$, where $\dot{X}_{j}$ is the $1$-order difference of $X_j$. 

The rate of $X_i$ marginal entropy change includes its own entropy change $ \frac{d H_i^*}{dt}$, the influence of noise $\frac{d H_i^{\text{noise}}}{dt} $ and the information flow from other time series $T_{j\to i}$. To measure the relative magnitude of causality, the information flow is normalized as:
\begin{align}
    Z = \left| \frac{d H_i^*}{dt} \right|& + \sum_{j\neq i} \left| T_{j \to i} \right| + \left| \frac{d H_i^{\text{noise}}}{dt} \right|,\\
    &\tau_{j \to i} = \frac{T_{j \to i}}{Z}.
\end{align}

EEG signals are usually collected by multi-channel BCI devices under the stimuli, and can be regarded as a group of time series. Therefore, the information flow between channels can be quantitatively calculated from EEG data by formula \ref{equ:1}, and the information flow can be normalized and the edges with significant causal relationship can be filtered through hypothesis testing, so as to construct causality graph $\mathcal{G}=(\mathbf{V},\mathbf{A},\mathbf{X})$. $\mathbf{V}$ is a set of $n$ nodes (electrode channels) and $\mathbf{X}\in \mathbb{R}^{n\times d}$ is the feature matrix. $\mathbf{A}\in \mathbb{R}^{n\times n}$ is the adjacent matrix whose elements $\mathbf{A}_{ij}$ represent the strength of the connection from $X_i$ to $X_j$. 
\begin{align}
    \mathbf{A_{ij}}=|\tau _{i\to j}|.
\end{align}

To mitigate the interference of noisy connections on model learning, $\mathbf{A}$ is sparsified by hypothesis testing of the information flow at the significance level of $\alpha=0.01$, and only statistically significant causal connections are retained.

\subsection{Preprocessing} 
The MEEG dataset\cite{xiao2024meeg} was collected using a 32-channel BCI acquisition device at a sampling rate of 1000 Hz, where EEG signals were recorded under 20 music stimuli, each lasting 1 min. During preprocessing, the raw EEG signals were first downsampled to 200 Hz, and then segmented into non-overlapping windows of 4s. Each segmented EEG sample was subsequently processed in two parallel ways. On the one hand, the EEG signals were treated as multivariate time series and used for causal matrix $A_{global}$ construction following the procedure described in Section \ref{sec:2.1}. On the other hand, we decompose EEG signals into five frequency bands($\delta$, $\theta$, $\alpha$, $\beta$, $\gamma$), and extract Differential Entropy (DE) as the node feature $X_{global}$.

\begin{figure}[ht]
    \centering
    \includegraphics[width=\linewidth]{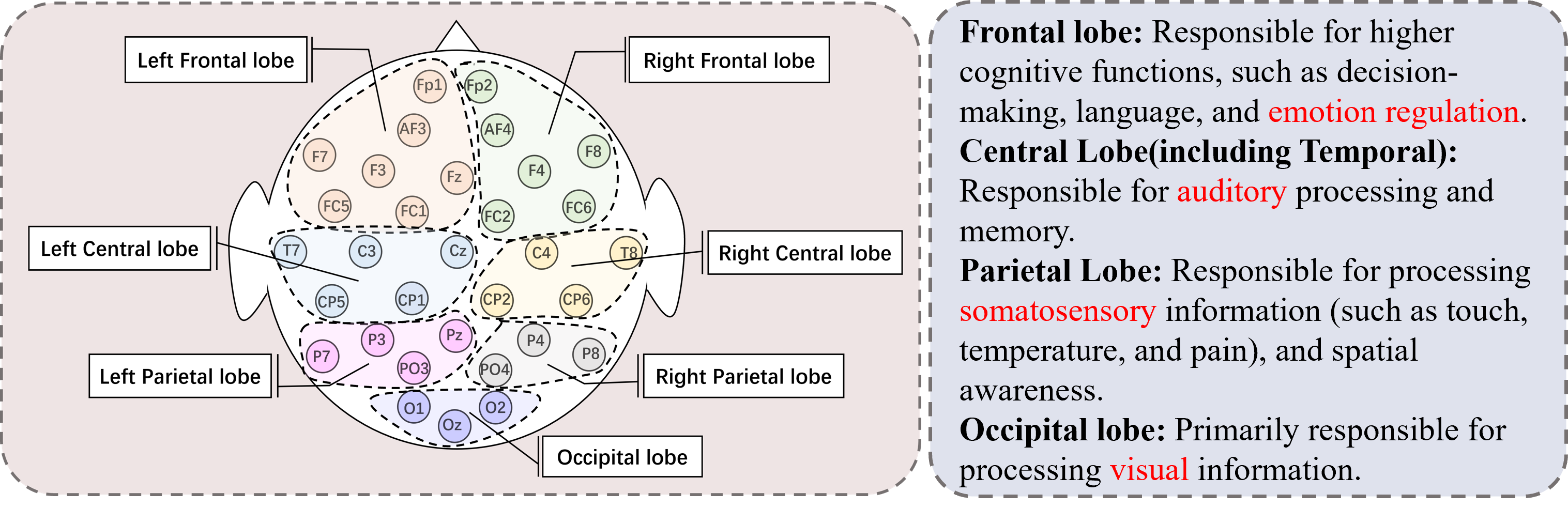}
    \caption{Brain regionalization}
    \label{fig:2}
    \vspace{-10pt}
\end{figure}

To capture information over short distances between channels, we construct a parallel local branch. Taking into account the asymmetry of brain regions and hemispheres, we divided EEG channels into seven regions as shown in the Fig.\ref{fig:2}.We reorder the EEG channels according to the above seven regions. The channels within each region are next to each other so that local causal matrix construction and aggregation operations can be performed by region. So $A_{local}$ is a block-diagonal matrix. In other words, information exchange occurs only within each region in local branch.
\subsection{Information aggregation}
We constructs Hierarchical Causality Graph Convolutional Neural Network (HCD -GCN) to complete information aggregation, which consists of two directed graph diffusion convolutional (DConv~\cite{li2018dcrnn}) layers and a graph pooling (DIFFPOOL~\cite{ying2018diffpool} or Attention Pool) layer.Assuming that the adjacency matrix is $A\in \mathbb{R}^{n\times n}$ and the feature matrix $X\in \mathbb{R}^{n\times d}$, the out-degree and in-degree are distinguished in a directed graph, and they are represented by $D_O=diag(\sum_{j=1}^{n} A_{1j}, \sum_{j=1}^{n} A_{2j}, \dots,$ $\sum_{j=1}^{n} A_{nj})$, $D_I = diag\left(\sum_{i=1}^{n} A_{i1}, \sum_{i=1}^{n} A_{i2}, \dots, \sum_{i=1}^{n} A_{in}\right).$

DConv considers both forward and backward diffusion processes to capture upstream and downstream information of nodes.Assuming $\Theta_{k,1}^l$ and $\Theta_{k,2}^l$ are the learnable parameter tensors, the q-dimensional features $X^{l+1}\in \mathbb{R}^{n\times q}$ of the $(l+1)$-th layer can be obtained from the d-dimensional features $X^{l}\in \mathbb{R}^{n\times d}$ of the $l$-th layer through DConv mapping:

\vspace{-10pt}
\begin{align}
    X^{l+1} = \sigma\Bigg[ \sum_{k=0}^{K-1} \left( \left( D_O^{-1} A \right)^k X^l \Theta_{k,1}^l + \left( D_I^{-1} A^\top \right)^k X^l \Theta_{k,2}^l \right) \Bigg].
\end{align}
\vspace{-10pt}

In the global branch, we set up two 4th-order DConv layers, and then use DIFFPOOL to automatically learn the hierarchical structure of the graph and coarsen the nodes information.
DIFFPOOL completes the feature mapping between layers through a learnable assignment matrix $S^{l}\in \mathbb{R}^{n_l\times n_{l+1}}$.The process is as follows:
\begin{align}
    S^l=softmax(GNN&_{l,pool}(X^l,A^l)),\nonumber\\
    Z^l=DConv(X^l,A^l)&, \, X^{l+1}= S^{(l)^\top} Z^{l}.
\end{align}

In the local branch, we set up two 2-order DConv layers due to the small number of nodes in each region, and then construct an intra-region Attention Pool layer to coarsen node information. Inspired by the Squeeze-and-Excitation Network (SENet), we employ a learnable subnetwork (MLP) to dynamically compute the weights factor $\mathbf{w}_k$ of all channels within the $k$-th region based on the its overall features, thereby enhancing important channel information.  The computation steps of Attention Pool are as follows:
\begin{align}
    \mathbf{w}&_k = f_{\text{SE}, k}(mean(X_k^l,dim=1)),\nonumber\\
    y_k = \sum_{i=1}^{C_k} & \mathbf{w}_{k, i} \cdot X_{k, i}^l, \, X_{l+1}=concat(y_k,dim=1).
\end{align}
where $X_k^l\in{R^{C_k\times d}}$ represents the feature matrix of the $k$-th region in the $l$-th layer, and $C_k$ is the number of channels in the $k$-th region.

\subsection{Branch fusion and classification}

After the dual-branch information aggregation module, we obtain global node representations $Z_{global}$ and local node representations $Z_{local}$, both of which contain seven abstract node features. To fully utilize the two types of features, we introduce a lightweight gated network that adaptively learns the importance weights of each abstract node in the global and local branches based on the initial input features, and performs weighted fusion of the outputs from the two branches along the feature dimension. The final joint representation $Z\in\mathbb{R}^{1\times hidden-dim}$ is then fed into the classifier composed of two fully connected layers with ReLU activation and dropout.

\section{Experiments}
\subsection{Implementation details}
All experiments are carried out under the cross-trail paradigm, using nested cross-validation, with 10-fold cross-validation in the outer loop and 3-fold cross-validation in the inner loop.The inner cross-validation adopts a two-stage training strategy. In the first stage, the candidate model with the best performance is selected using a learning rate of $10^{-3}$ and 200 epochs. In the second stage, the selected model is further optimized on the entire training set with a learning rate of $10^{-4}$ for fine-tuning, and epochs is set to 20. All training procedures use the Adam optimizer with a batch size of 64.

\subsection{Comparison with SOTA Methods}
Table~\ref{tab:6.1} benchmarks GL-LFGNN against CNN-based (EEGNet~\cite{lawhern2018eegnet}, EEG-TCNet~\cite{ingolfsson2020eegtcnet}, ATCNet~\cite{altaheri2023atcnet}), GNN-based (DGCNN~\cite{song2020dgnn}, LGGNet~\cite{ding2024lggnet}, AT-DGNN~\cite{xiao2024meeg}), and transformer-based (EmT~\cite{ding2025emt}) methods. Three key observations emerge: \textbf{(1) Arousal advantage:} GL-LFGNN achieves the highest arousal accuracy (86.17\%) and notably the \textit{lowest variance} (std=10.10), suggesting that causal graph structure provides more stable cross-subject generalization---consistent with evidence that arousal involves robust subcortical-cortical pathways amenable to causal modeling. \textbf{(2) Valence competitiveness:} While EmT leads on valence (87.96\% vs 86.71\%), GL-LFGNN attains superior consistency (std=8.87 vs 12.49), reflecting reduced overfitting from explicit structural priors. \textbf{(3) Efficiency:} With only 37K parameters ($\sim$10\% of EmT, $\sim$1.4\% of AT-DGNN), GL-LFGNN demonstrates that encoding domain knowledge via causal graphs substantially reduces model complexity without sacrificing accuracy.

\begin{table}[ht] 
\vspace{-10pt}
    \centering
    \caption{Compare with other methods on MEEG dataset.The data in the table represent the "mean/std", with the \textcolor{blue}{optimal values} for each metric highlighted in blue and the \textcolor{purple}{suboptimal values} highlighted in purple.}
    \label{tab:6.1}

    \resizebox{\textwidth}{!}{
    \begin{tabular}{l|l|cc|cc|c}
        \hline
        \multirow{2}{*}{\textbf{Method}} & \multirow{2}{*}{\textbf{Year}} & 
        \multicolumn{2}{c|}{\textbf{Arousal}} & 
        \multicolumn{2}{c|}{\textbf{Valence}} &  
        \multirow{2}{*}{\textbf{\#params}}\\
            \cline{3-6}
            & & ACC (\%)  & F1 (\%)  & ACC (\%) & F1 (\%) &\\
            \hline
            EEGNet\cite{lawhern2018eegnet} & 2018 & $75.47 / 17.63$ & $75.54/17.83$ & $73.66/16.82$ & $73.33 / 16.94$ &\textcolor{purple}{14,706} \\
            \hline
            EEG-TCNet\cite{ingolfsson2020eegtcnet}& 2020 & $76.36 /17.05 $& $73.79/19.66$& $69.72/16.06$ & $54.28/22.43$ & \textcolor{blue}{4,374}\\
            \hline
            DGCNN\cite{song2020dgnn}& 2020 & $82.12 / 12.05$ & $81.83 / 12.45$ & $82.72 / 11.00$ & $82.40 / 11.44$ & 68,420 \\
            \hline
            ATCNet\cite{altaheri2023atcnet}& 2023 & $82.01/ 14.06$ & $79.17/17.57$ & $83.19 / 12.09$ & $82.08 / 13.41$ &575,570\\
            \hline
            LGGNet-Fro\cite{ding2024lggnet}& 2024 & $81.85/ 12.57$ & $81.58 /12.73$ & $84.28 / 11.83$ & $83.87 /12.66$ & 1,135,230 \\
            LGGNet-Gen& & $82.15/ 12.72$ & $82.15 / 12.57$ & $84.53/ 11.44$ & $84.38 / 11.56$  &\\
            LGGNet-Hem& & $81.92 / 12.08$ & $81.58 / 12.43$ & $84.93/ 11.21$ & $84.46 / 12.27$  & \\
            \hline
            AT-DGNN-Fro\cite{xiao2024meeg}&2024 & $83.51 / $ \textcolor{purple}{11.09}& $83.06 /11.83$ & $85.56 /10.69 $& $85.61 / 10.89$ & 2,680,350\\
            AT-DGNN-Gen& & $83.74 /12.03 $& $84.56 /$ \textcolor{purple}{10.88} & $86.01 / $\textcolor{purple}{9.40} & $85.68 /$ \textcolor{purple}{10.06} & \\
            AT-DGNN-Hem& & $83.73 / 11.20 $& $84.78 / 10.93$ & $84.89/ 10.57$ & $85.35 / 10.24$ &\\
            \hline
            EmT\cite{ding2025emt}&2025& \textcolor{purple}{85.31}/14.30 & \textcolor{purple}{85.40}/ 14.10  & \textcolor{blue}{87.96}/12.49 &  \textcolor{blue}{87.94}/12.14 & 352,738\\
            \hline
            \textbf{GL-LFGNN}& & \textcolor{blue}{86.17}/\textcolor{blue}{10.10}&\textcolor{blue}{86.45}/\textcolor{blue}{9.80} & \textcolor{purple}{86.71}/\textcolor{blue}{8.87 } & \textcolor{purple}{86.53}/\textcolor{blue}{9.42} &37,166\\
            \hline
        \end{tabular}
    }
\end{table}

\vspace{-20pt}
\subsection{Ablation experiments}
We validate core design choices through systematic ablation (Table~\ref{tab:1}). \textbf{Dual-branch synergy:} The full model (86.17\%/86.71\%) significantly outperforms single branches (paired Wilcoxon test, $p<0.05$), with improvements of +2.7\%/+2.2\% over global-only and +2.8\%/+4.5\% over local-only configurations. This synergy aligns with neuroscientific understanding: emotion processing involves both distributed long-range connectivity (captured by global branch) and localized computations within specialized regions (local branch). Notably, the local branch alone underperforms on valence (82.21\%), consistent with findings that valence discrimination requires integration across hemispheres---a capacity restored by global-local fusion.

\vspace{-10pt}
\begin{table}[ht]
    \centering 
    \caption{ Ablation experiments results. "Global" and "Local" represent retaining only the corresponding single branch.}
    \label{tab:1}
    \resizebox{0.9\textwidth}{!}{
    \begin{tabular}{|c|c|c|c|c|}
    \hline
    \multirow{2}{*}{Branch} 
    & \multicolumn{2}{c|}{Arousal(\%)} 
    & \multicolumn{2}{c|}{Valence(\%)} \\ \cline{2-5}
    & Accuracy &F1-score 
    & Accuracy & F1-score \\ \hline
    Global
    & $83.46 \pm 11.82$ & $83.51 \pm 11.86$  & $84.53\pm 10.16$ & $84.45 \pm 10.34 $ \\ \hline
    Local
    & $83.37 \pm 10.58$ & $83.85 \pm 10.02$  & $82.21 \pm 11.02$ & $81.73 \pm 12.05$ \\ \hline
    \textbf{Proposed}
    & \textbf{86.17 $\pm$ 10.10} &\textbf{86.45 $\pm$ 9.80} &  \textbf{86.71 $\pm$ 8.87}  & \textbf{86.53 $\pm$ 9.42}\\ \hline
    \end{tabular}
    }
\end{table}
\vspace{-10pt}

\textbf{Causal method comparison:} Table~\ref{tab:2} contrasts Liang-Kleeman vs Granger causality graphs (both with Top-k sparsification) on Subject 25. Liang-Kleeman yields consistent gains (+0.71\% arousal, +2.50\% valence), with the larger valence improvement suggesting its advantage in capturing the subtle, multi-directional causal flows underlying complex affective states. This supports our hypothesis: Granger causality's reliance on autoregressive prediction conflates temporal precedence with causation, while Liang-Kleeman's entropy-based formulation isolates genuine information transfer---particularly valuable for non-stationary emotional dynamics.

\vspace{-10pt}
\begin{table}[htbp]
    \centering 
    \caption{ Case Study: performance of different graph construction methods (sub 25)}
    \label{tab:2}
    \resizebox{0.7\textwidth}{!}{
    \begin{tabular}{|c|c|c|c|c|}
    \hline
    \multirow{2}{*}{Causal Graph} 
    & \multicolumn{2}{c|}{Arousal(\%)} 
    & \multicolumn{2}{c|}{Valence(\%)} \\ \cline{2-5}
    & Accuracy &F1-score 
    & Accuracy & F1-score \\ \hline
    Granger 
    & $93.93$ & $93.95$  & $85.71$ & $86.49 $ \\ \hline
    \textbf{Liang-Kleeman}
    & \textbf{94.64} & \textbf{94.77}  & \textbf{88.21}& \textbf{88.17} \\ \hline
    \end{tabular}
    }
\end{table}
\vspace{-10pt}

\begin{figure}[ht]
\vspace{-10pt}
  \centering
    \includegraphics[width=\linewidth]{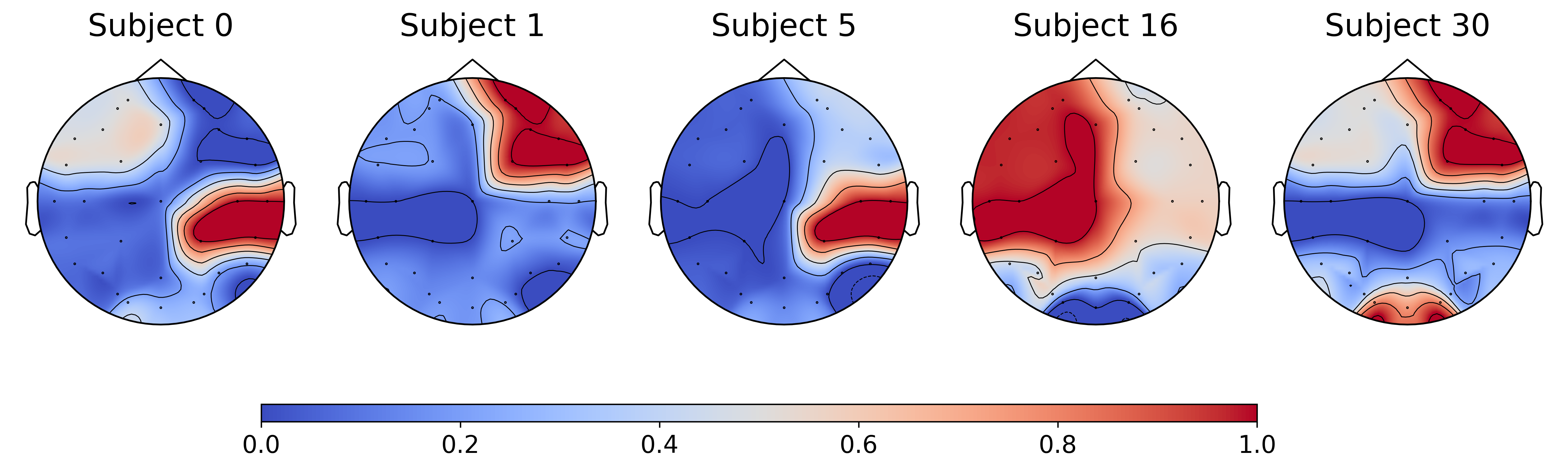}
  \caption{Visualization of local branch attention weights witch are the average of 10-fold cross-validation.}
  \label{fig:3}
\end{figure}

\textbf{Regional importance analysis:} Visualization of local branch attention weights (Fig.\ref{fig:3}) reveals that Frontal and Central-Temporal regions (Fig.\ref{fig:2}) contribute most to arousal classification. This pattern is neurophysiologically coherent: (1) the prefrontal cortex orchestrates top-down emotion regulation and arousal modulation via connections to the amygdala; (2) the auditory cortex in temporal regions directly processes music stimuli that induce emotional responses in the MEEG paradigm. Such interpretable attention patterns---emergent from causal graph learning rather than explicit supervision---validate that GL-LFGNN captures functionally meaningful brain organization.

\vspace{-25pt}
\section{Conclusion}
We present GL-LFGNN, which bridges a critical gap between correlational graph learning and the causal nature of neural information flow. By grounding adjacency matrices in Liang-Kleeman information flow theory, our model provides both rigorous causal semantics and neurophysiological interpretability---regional attention patterns align with established emotion circuitry without explicit supervision. Notably, GL-LFGNN achieves SOTA-comparable performance using only $\sim$10\% of the parameters required by leading methods, suggesting that principled causal structure can substitute for architectural complexity---a promising direction for deploying interpretable affective computing in resource-constrained wearable and clinical settings. Future work will extend this framework to cross-dataset generalization and real-time affective monitoring applications.

%
%
%

\end{document}